\documentclass[sigconf]{acmart}

\AtBeginDocument{%
  }

\newcommand{\bc}[1]{\left\{{#1}\right\}}

\newcommand{\bs}[1]{\left[{#1}\right]}

\newcommand{\abs}[1]{\left\vert#1\right\vert}

\newcommand{\prob}[1]{\mathbb{P}\bs{#1}}

\usepackage{subcaption}

\begin{document}

\title[Evaluating Fairness in Transaction Fraud Models]{Evaluating Fairness in Transaction Fraud Models: Fairness Metrics, Bias Audits, and Challenges}

%\author{Featurespace Innovation Lab}
%\affiliation{
%  \institution{Featurespace}
%  \city{Cambridge}
%  \country{UK}}
%\email{innovation@featurespace.co.uk}

\author{Parameswaran Kamalaruban}
\affiliation{
 \department{Innovation Lab}
 \institution{Featurespace}
 \city{Cambridge}
 \country{UK}}
\email{kamal.parameswaran@featurespace.co.uk}

\author{Yulu Pi}
\affiliation{
 \department{Centre for Interdisciplinary Methodologies}
 \institution{University of Warwick}
 \city{Coventry}
 \country{UK}}
\email{yulu.pi@warwick.ac.uk}

\author{Stuart Burrell}
%\authornote{Note}
%\orcid{0000-0002-6333-1750}
\affiliation{
  \department{Innovation Lab}
  \institution{Featurespace}
  \city{Cambridge} 
  \country{UK} 
}
\email{stuart.burrell@featurespace.co.uk}

\author{Eleanor Drage}
\affiliation{
 \department{The Leverhulme Centre for the Future of Intelligence}
 \institution{University of Cambridge}
 \city{Cambridge}
 \country{UK}}
\email{ed575@cam.ac.uk}

\author{Piotr Skalski}
%\authornote{Note}
%\orcid{0000-0003-3102-9837}
\affiliation{
  \department{Innovation Lab}
  \institution{Featurespace}
  \city{Cambridge} 
  \country{UK} 
}
\email{piotr.skalski@featurespace.co.uk}

\author{Jason Wong}
%\authornote{Note}
%\orcid{0000-0001-7727-1341}
\affiliation{
  \department{Innovation Lab}
  \institution{Featurespace}
  \city{Cambridge} 
  \country{UK}
}
\email{jason.wong@featurespace.co.uk}

\author{David Sutton}
%\authornote{Note}
%\orcid{0009-0005-6739-5689}
\affiliation{
  \department{Innovation Lab}
  \institution{Featurespace}
  \city{Cambridge} 
  \country{UK} 
}
\email{david.sutton@featurespace.co.uk}

%\renewcommand{\shortauthors}{Innovation et al.}

% !TEX root =  main.tex
%%%%%%%%%%%%%%%%%%%%%%%%%%%%%%%%%%%%%
%%%%%%%%%%%%%%%%%%%%%%%%%%%%%%%%%%%%%
\begin{abstract}
Ensuring fairness in transaction fraud detection models is vital due to the potential harms and legal implications of biased decision-making. Despite extensive research on algorithmic fairness, there is a notable gap in the study of bias in fraud detection models, mainly due to the field's unique challenges. These challenges include the need for fairness metrics that account for fraud data's imbalanced nature and the tradeoff between fraud protection and service quality. To address this gap, we present a comprehensive fairness evaluation of transaction fraud models using public synthetic datasets, marking the first algorithmic bias audit in this domain. Our findings reveal three critical insights: (1) Certain fairness metrics expose significant bias only after normalization, highlighting the impact of class imbalance. (2) Bias is significant in both service quality-related parity metrics and fraud protection-related parity metrics. (3) The fairness through unawareness approach, which involved removing sensitive attributes such as gender, does not improve bias mitigation within these datasets, likely due to the presence of correlated proxies. We also discuss socio-technical fairness-related challenges in transaction fraud models. These insights underscore the need for a nuanced approach to fairness in fraud detection, balancing protection and service quality, and moving beyond simple bias mitigation strategies. Future work must focus on refining fairness metrics and developing methods tailored to the unique complexities of the transaction fraud domain.
\end{abstract}




%\received{2 August 2024}
%\received[revised]{2 August 2024}
%\received[accepted]{2 August 2024}

\maketitle

% !TEX root =  main.tex
%%%%%%%%%%%%%%%%%%%%%%%%%%%%%%%%%%%%%
%%%%%%%%%%%%%%%%%%%%%%%%%%%%%%%%%%%%%

\section{Introduction}
\label{sec:introduction}

The rise of digital transactions has brought about significant advancements in convenience and efficiency. However, this growth has also been accompanied by a surge in transaction fraud, necessitating the development of robust fraud detection models~\cite{jha2012employing,afriyie2023supervised,skalski2023towards}. While these models have evolved to become more sophisticated, an equally important concern has emerged: ensuring fairness in their predictions and in how these predictions are utilized within the wider decision-making system. The potential harms of biased fraud detection systems are profound, ranging from unfair denial of services to specific demographic groups to severe legal repercussions for financial institutions. Despite extensive research on algorithmic fairness in various domains, a notable gap exists in understanding and addressing bias within transaction fraud detection models.

Ensuring fairness in transaction fraud models is not just a technical necessity but also a regulatory imperative. Regulatory frameworks like the US Civil Rights Act of 1964 and the EU's General Data Protection Regulation (GDPR) mandate unbiased decision-making. Additional regulations in the finance industry, such as the Fair Credit Reporting Act (FCRA), the Equal Credit Opportunity Act (ECOA), and SR 11-7: Guidance on Model Risk Management, further emphasize the need for fairness in machine learning (ML) models deployed in financial contexts~\cite{das2021fairness}. Recent guidelines, including the FDIC Model Risk Guidance (2017), Article 4 of the Technical Guidance of FCA (2019), the US Treasury Compliance Framework (2023), and the UK ISO Guidance on AI and Data Protection (2023), further highlight the evolving landscape of compliance requirements. Meeting these regulatory requirements begins with accurately measuring bias in ML models, which is the first step toward mitigating bias and achieving compliance.

The importance of fairness-aware machine learning extends beyond finance, encompassing areas such as hiring, criminal justice, medicine, and college admissions. For instance, a recruitment tool used by Amazon was shown to be biased against women, highlighting the pervasive nature of this issue across different domains~\cite{dastin2022amazon}. A system is considered fair if it does not discriminate based on protected characteristics such as race, sex, or religion. However, deployed ML classifiers often demonstrate bias towards certain demographic groups present in the data~\cite{dwork2012fairness}, which has the potential to exacerbate existing social inequities~\cite{o2017weapons}. Therefore, the need for fair ML models is critical~\cite{mehrabi2021survey,caton2024fairness}.

Transaction fraud detection models face unique challenges that distinguish them from other areas of machine learning. One primary challenge is the inherently imbalanced nature of fraud data, where fraudulent transactions constitute a minute fraction of the total transaction volume~\cite{he2009learning}. This imbalance complicates the application of standard fairness metrics, which may fail to account for the disproportionate representation of classes. Furthermore, transaction data often exhibit sequential characteristics, adding another layer of complexity in measuring and ensuring fairness. These aspects necessitate developing and applying specialized fairness metrics tailored to the context of transaction fraud detection.

To bridge this gap, we conducted a systematic bias audit on transaction fraud detection models using public transaction fraud datasets. The models used in our study are LightGBM classifiers trained with features suggested by~\cite{jha2012employing}. Our audit focused on evaluating the performance of these models through the lens of fairness, employing various group fairness metrics. We also explored the effectiveness of the fairness through unawareness approach, which involves removing sensitive attributes such as gender from the models. Our findings underscore the limitations of this approach, especially when proxies correlated with the protected attributes are present. In such cases, biases persist even after the protected attributes are removed.

In this paper, we make the following contributions:
\begin{enumerate}
\item We provide a comprehensive discussion of group fairness metrics in the context of transaction fraud models and categorize these metrics based on their objectives (Section~\ref{subsec:metrics}).
\item We highlight the challenges in selecting appropriate fairness metrics, normalizing them to address the class imbalance, and accounting for the transaction value (Section~\ref{subsec:algo-challenges}).
\item We present empirical findings from a bias audit conducted on public transaction datasets, demonstrating significant biases in service quality and fraud protection metrics (Section~\ref{sec:experiments}).
\item We discuss the socio-technical challenges of formalizing and operationalizing fairness in transaction fraud detection, emphasizing the need for nuanced approaches that balance protection and service quality (Section~\ref{sec:socio-tech-fairness}).
\end{enumerate}
These contributions aim to advance the understanding of fairness in transaction fraud detection and provide a foundation for developing more equitable models in this critical domain.

% !TEX root =  main.tex
%%%%%%%%%%%%%%%%%%%%%%%%%%%%%%%%%%%%%
%%%%%%%%%%%%%%%%%%%%%%%%%%%%%%%%%%%%%

\section{Related Work}
\label{sec:relatedwork}

\paragraph{Algorithmic Fairness} Algorithmic fairness is a rapidly evolving field focused on ensuring equitable outcomes in AI decision-making systems. Various approaches have been proposed to measure and mitigate biases within algorithms~\cite{verma2018fairness,bellamy2019ai,weerts2023fairlearn,barocas2023fairness}. However, these bias mitigation methods often grapple with trade-offs between fairness and other objectives, such as accuracy, highlighting the ongoing debate about optimal fairness definitions and their practical implementation~\cite{kleinberg2016inherent}. The complexity of balancing multiple fairness metrics is exacerbated by their inherent conflicts, where improving fairness on one metric can degrade it on another~\cite{kim2020fact,berk2021fairness}.

\paragraph{Socio-Technical Fairness} Socio-technical fairness extends the discussion of algorithmic fairness into the broader social and organizational contexts in which these systems operate~\cite{dolata2022sociotechnical,lee2021formalising,so2022beyond,smith2024recommend}. This perspective acknowledges that biases in AI can reflect and amplify existing societal inequalities and that fairness cannot be fully achieved without considering these contextual factors~\cite{selbst2019fairness,o2017weapons}. It emphasizes the importance of involving diverse stakeholders in designing and deploying AI systems to address systemic issues and ensure that fairness interventions are contextually relevant. The literature highlights that socio-technical considerations can significantly impact the effectiveness of fairness measures and advocates for a holistic approach integrating technical solutions with regulatory and organizational reforms~\cite{barocas2016big}.

\paragraph{Fairness in Financial AI} While algorithmic fairness has been extensively studied in various domains, its application to finance is still emerging. Studies have explored bias in areas such as credit scoring~\cite{kozodoi2022fairness}, bank account-opening fraud detection~\cite{pombal2022understanding}, and lending decision making~\cite{alesina2013women}. Biased models can disproportionately harm certain groups, limiting their access to financial services, economic opportunities, and individual livelihoods~\cite{barocas2016big,bartlett2022consumer}. However, there are still specific challenges and deficiencies in achieving fairness within the financial industry. Research has highlighted the challenges of implementing fairness-aware ML pipelines in financial services and the practical difficulties of applying fairness techniques, particularly in the context of loans~\cite{das2021fairness}. Transaction fraud detection presents its own unique challenges within the financial industry, necessitating further exploration and tailored solutions to address these issues.

% !TEX root =  main.tex
%%%%%%%%%%%%%%%%%%%%%%%%%%%%%%%%%%%%%
%%%%%%%%%%%%%%%%%%%%%%%%%%%%%%%%%%%%%
\section{Algorithmic Fairness in Transaction Fraud Models} 
\label{sec:algo-fairness}

In this section, we examine the concept of algorithmic fairness as applied to transaction fraud detection models, focusing on a range of group fairness metrics. We categorize these metrics based on their objectives--either ensuring equal protection against fraud or delivering uniform quality of service across diverse demographic groups. Additionally, we delve into the complexities involved in the algorithmic bias audit process, addressing challenges such as class imbalance, the sequential nature of transaction data, and the importance of transaction value. This discussion sets the stage for understanding how fairness can be effectively evaluated in transaction fraud models.

\subsection{Fairness Metrics}
\label{subsec:metrics}

Given a wide array of group fairness metrics proposed in the literature and the disagreement between these metrics, selecting an appropriate metric is problem-dependent. In the following, we discuss several group fairness metrics in the context of transaction fraud models, considering a binary sensitive attribute $G \in \{ \textnormal{Male}, \textnormal{Female} \}$.

\subsubsection{Classifier threshold dependent metrics.} First, we discuss group fairness metrics based on binary predictions from the classifier, which necessitate selecting a threshold value for the classifier's predictions.  

\paragraph{Recall Parity or Equal Opportunity~\cite{hardt2016equality}.} Recall or True Positive Rate (\texttt{TPR}) is the ratio of true positives to the total actual positives, i.e., $\frac{\texttt{TP}}{\texttt{TP} + \texttt{FN}}$. This metric measures the absolute difference in recall between genders: $\abs{\texttt{TPR}_{G=\textnormal{Male}} - \texttt{TPR}_{G=\textnormal{Female}}}$. A high value indicates that the model detects fraud better for one gender, indicating potential bias.

\paragraph{Negative Predictive Value Parity.} Negative Predictive Value (\texttt{NPV}) measures the ratio of true negatives to the total predicted negatives, i.e., $\frac{\texttt{TN}}{\texttt{TN} + \texttt{FN}}$. This metric measures the absolute difference in \texttt{NPV} between genders: $\abs{\texttt{NPV}_{G=\textnormal{Male}} - \texttt{NPV}_{G=\textnormal{Female}}}$. A high value indicates that the model is more likely to be correct about non-fraudulent transaction predictions for one gender, suggesting bias.  

\paragraph{Precision Parity~\cite{chouldechova2017fair}.} Precision or Positive Predictive Value (\texttt{PPV}) is the ratio of true positives to predicted positives, i.e., $\frac{\texttt{TP}}{\texttt{TP} + \texttt{FP}}$. This metric measures the absolute difference in precision between genders: $\abs{\texttt{PPV}_{G=\textnormal{Male}} - \texttt{PPV}_{G=\textnormal{Female}}}$. A high value indicates that the model is more accurate in predicting fraud for one gender than the other, suggesting bias.  

\paragraph{False Positive Rate Parity~\cite{chouldechova2017fair}.} False Positive Rate (\texttt{FPR}) is the ratio of false positives to the total actual negatives, i.e., $\frac{\texttt{FP}}{\texttt{FP} + \texttt{TN}}$. This metric measures the absolute difference in \texttt{FPR} between genders: $\abs{\texttt{FPR}_{G=\textnormal{Male}} - \texttt{FPR}_{G=\textnormal{Female}}}$. A high value indicates that the model's rate of false alarms differs between genders, suggesting bias. 

\paragraph{F1 Score Parity.} The F1 score (\texttt{F1}) is a measure that combines both precision and recall. This metric measures the absolute difference in F1 scores between genders: $\abs{\texttt{F1}_{G=\textnormal{Male}} - \texttt{F1}_{G=\textnormal{Female}}}$. A high value indicates that the model has different precision and recall balances for each gender, suggesting bias in how well the model predicts fraudulent transactions for different genders. 

\paragraph{Equalized Odds~\cite{hardt2016equality}.} This measures the absolute differences between genders in both \texttt{TPR} and \texttt{FPR}: 
\[
\abs{\texttt{TPR}_{G=\textnormal{Male}} - \texttt{TPR}_{G=\textnormal{Female}}} + \abs{\texttt{FPR}_{G=\textnormal{Male}} - \texttt{FPR}_{G=\textnormal{Female}}} .
\]
A high value indicates significant disparities in either \texttt{TPR} or \texttt{FPR}, suggesting bias. 

\paragraph{Demographic Parity~\cite{dwork2012fairness}.} This measures the absolute difference in the predicted fraud rates for males and females: 
\[
\abs{\prob{\widehat{Y} = 1 \mid G = \textnormal{Male}} - \prob{\widehat{Y} = 1 \mid G = \textnormal{Female}}} ,
\]
where $\widehat{Y}$ is the model prediction. A high value indicates different predicted fraud rates for each gender, suggesting bias.

\subsubsection{Classifier threshold independent metrics.} Now, we discuss group fairness metrics based directly on the classifier's probabilistic predictions.  

\paragraph{ROC AUC Parity.} The ROC AUC score (\texttt{ROC-AUC}) measures the model's ability to distinguish between classes. This metric measures the absolute difference in \texttt{ROC-AUC} between genders: 
\[
\abs{\texttt{ROC-AUC}_{G=\textnormal{Male}} - \texttt{ROC-AUC}_{G=\textnormal{Female}}} .
\]
A high value indicates different discrimination capabilities for each gender, suggesting bias.

\paragraph{PR AUC Parity.} The Precision-Recall AUC score (\texttt{PR-AUC}) measures the relationship between precision and recall across different thresholds. This metric measures the absolute difference in \texttt{PR-AUC} between genders: $\abs{\texttt{PR-AUC}_{G=\textnormal{Male}} - \texttt{PR-AUC}_{G=\textnormal{Female}}}$. A high value indicates that the model's ability to identify fraudulent transactions varies significantly between genders, suggesting potential bias. 

\paragraph{TPR Parity at different FP-ratio thresholds.} The false positive ratio (\texttt{FP-ratio}) is the number of false positives compared to the number of true positives. \texttt{TPR @ FP-ratio} assesses the model's ability to correctly identify fraudulent transactions for a given \texttt{FP-ratio}. This metric measures the absolute difference in \texttt{TPR} between genders at specified \texttt{FP-ratio} thresholds: 
\[
\abs{\texttt{TPR @ FP-ratio}_{G=\textnormal{Male}} - \texttt{TPR @ FP-ratio}_{G=\textnormal{Female}}} .
\]
A high value indicates that the model's sensitivity to fraud detection varies significantly between genders at these false positive ratios, suggesting potential bias.

\paragraph{VDR Parity at different FP-ratio thresholds.} The Value Detection Rate (\texttt{VDR}) is the true positive rate weighed by transaction value. \texttt{VDR @ FP-ratio} evaluates the model's effectiveness in identifying high-value fraudulent transactions for a given \texttt{FP-ratio}. This metric measures the absolute difference in \texttt{VDR} between genders at specified \texttt{FP-ratio} thresholds: 
\[
\abs{\texttt{VDR @ FP-ratio}_{G=\textnormal{Male}} - \texttt{VDR @ FP-ratio}_{G=\textnormal{Female}}} .
\]
It is important to analyze \texttt{VDR} parity alongside \texttt{TPR} parity. If \texttt{TPR} parity is low but \texttt{VDR} parity is high, it indicates a specific bias in detecting high-value fraud. This combined analysis clarifies whether the bias is more pronounced in high-value transactions.

\subsection{Fairness Metrics Grouping} 

Transaction fraud models are often judged by their ability to reduce fraud losses while maintaining high precision score thresholds. False positive predictions result in declined transactions, leading to losses for the financial institution and negatively impacting the consumer experience. Thus, an appropriate performance metric should evaluate the volume or value of fraudulent transactions prevented against a certain threshold of declined genuine transactions. We group the above-discussed fairness metrics as follows:
\begin{itemize}
\item Fairness metrics that are related to the protection against fraud (fraud detection capability): Recall (TPR) Parity, VDR Parity, and NPV Parity.
\item Fairness metrics that are related to the quality of service (false alarm): Precision (PPV) Parity and FPR Parity.
\item Fairness metrics that are related to both groups: F1 Score Parity, Equalized Odds, Demographic Parity, ROC AUC Parity, and PR AUC Parity.
\item Fairness metrics that are related to the protection against fraud for a fixed quality of service: TPR or VDR Parity @ the FP ratio Threshold.
\end{itemize}

\subsection{Bias Audits}
\label{subsec:audits}

Given an evaluation dataset $(X,y)$ and a model $f: \mathcal{X} \to [0,1]$, we obtain the model predictions $\widehat{y}$ and $\widehat{y}_{\textnormal{binary}}$ for $X$ as follows: $\widehat{y} = f(X)$ denotes the class probability prediction, and $\widehat{y}_\textnormal{binary} = \widehat{y} \geq \tau$ denotes binary class prediction. The threshold $\tau$ should be carefully selected to account for class imbalance. For the transaction-level fairness evaluation, we group $(X,y)$ based on gender, i.e., $(X,y) = \cup_{g \in \mathcal{G}} (X_g, y_g)$, where $g \in \mathcal{G} = \bc{\textnormal{Male}, \textnormal{Female}}$ denotes gender. Then, for each $g \in \mathcal{G}$, we obtain the model predictions $\widehat{y}_g$ and $\widehat{y}_{\textnormal{binary},g}$ for $X_g$, as explained above. Finally, we use $y_g$, $\widehat{y}_g$, and $\widehat{y}_{\textnormal{binary},g}$ to compute the utility metrics (such as \texttt{PPV}, and \texttt{TPR}) for each $g \in \mathcal{G}$. These utility metrics are then used to compute fairness metrics, e.g., recall parity using $\abs{\texttt{TPR}_{g=\textnormal{Male}} - \texttt{TPR}_{g=\textnormal{Female}}}$.

\subsection{Challenges}
\label{subsec:algo-challenges}

\paragraph{Choice of Appropriate Fairness Metric.} In transaction fraud detection, where neither outcome (fraud or non-fraud) can be clearly labeled as favorable or unfavorable as in hiring or lending, choosing an appropriate group fairness metric should be done only after evaluating the impacts of accurately detecting fraud and avoiding incorrect fraud identifications that inconvenience customers. Metrics such as F1 Score Parity, Equalized Odds, and PR AUC Parity attempt to account for both quality of service harm (false alarm rate) and distributive harm (protection against fraud). Depending on the context, it may be advisable to consider metrics that balance these two types of harm with different weightings.

\paragraph{Class Imbalance Problem.} When dealing with imbalanced datasets, as is common in transaction fraud detection where fraud labels are rare compared to non-fraud labels, choosing group fairness metrics that appropriately handle this imbalance is essential. For all fairness metrics discussed in Section~\ref{subsec:metrics}, we also compute their normalized variants. For example, we compute the normalized FPR parity as follows:
\[
\frac{\abs{\texttt{FPR}_{G=\textnormal{Male}} - \texttt{FPR}_{G=\textnormal{Female}}}}{\max \bc{\texttt{FPR}_{G=\textnormal{Male}}, \texttt{FPR}_{G=\textnormal{Female}}}} .
\]
This normalization is essential for identifying bias due to the class imbalance inherent in the transaction fraud detection problem. For instance, in our experiments in Section~\ref{sec:experiments}, significant biases in important metrics such as Precision parity, FPR parity, and Equalized Odds go unnoticed without this normalization.

\paragraph{Sequential Nature of Transaction Data.} Transaction fraud detection datasets are inherently time-series in nature, composed of sequences of transactions over time. Each transaction is part of a continuous stream linked to an account, presenting unique challenges for fairness evaluation. In our experiments, we conduct transaction-level fairness evaluation, assuming that each account is associated with a single individual. However, in practice, accounts may have multiple authorized users, aggregate multiple cards, or be used by multiple people. Beyond transaction-level evaluation, assessing fairness at an aggregated or cardholder level is crucial, as the cumulative impact of multiple transactions affects the overall cardholder experience. We discuss the challenges of cardholder-level evaluation in Section~\ref{sec:socio-tech-fairness}.

%Transaction fraud detection datasets are inherently time-series in nature, presenting unique challenges for fairness evaluation. It may be appropriate to evaluate the quality of service harm-related fairness at an aggregated level (e.g., cardholder level) since the cumulative impact of multiple transactions influences the overall experience of the cardholder. However, distributive harm-related fairness should be assessed at the event level (e.g., transaction level) because every transaction needs to be individually protected to ensure that fraud detection mechanisms do not disproportionately impact certain demographic groups.

\paragraph{Importance of Transaction Value.} In transaction fraud detection, balancing the value of fraudulent and genuine transactions across demographic groups is crucial. This challenge is addressed by metrics like Value Detection Rate (VDR) Parity, which ensures the proportion of detected fraud value is equitable across groups. Focusing solely on transaction count may overlook significant financial disparities, as high-value fraud missed for certain groups can lead to severe consequences despite a balanced detection rate. VDR Parity provides a comprehensive measure of fairness by considering both transaction quantity and value, which is essential for equitable fraud detection outcomes. Moreover, the issue of liability--whether the financial institution or the consumer bears responsibility for the fraud--adds further complexity. Balancing these factors requires careful consideration of both statistical and economic impacts to ensure fair protection for all demographic groups against financial fraud.

\paragraph{Demographic Data for Fairness Evaluation.} Evaluating transaction fraud models with respect to group fairness metrics necessitates access to demographic information. Most fairness literature, encompassing fairness evaluation and bias mitigation techniques such as pre-processing, in-processing, and post-processing, assumes the availability of demographic data. However, obtaining demographic information in the context of transaction fraud detection presents significant challenges. Privacy concerns and regulatory constraints often restrict direct access to such data. In~\cite{chen2019fairness}, the authors employ a probabilistic predictor of protected/sensitive demographic attributes based on observable proxies, such as inferring race, ethnicity, or national origin via surname and geolocation. They compare the effectiveness of thresholded and weighted estimators of sensitive attributes in fairness evaluation. Techniques like leveraging large language models (LLMs) or weak supervision can also aid in sensitive attribute prediction. For a comprehensive overview, readers are referred to a recent survey on fairness without demographics~\cite{ashurst2023fairness}.

% For instance, in~\cite{zhu2023weak}, selected group fairness metrics are calibrated to remedy bias in proxy measures of sensitive attributes. Additionally, in~\cite{elzayn2023estimating}, fairness evaluation is considered in settings with limited access to protected attribute labels, assuming access to such labels for a small subset of the dataset and using probabilistic estimates for the rest. 

% !TEX root =  main.tex
%%%%%%%%%%%%%%%%%%%%%%%%%%%%%%%%%%%%%
%%%%%%%%%%%%%%%%%%%%%%%%%%%%%%%%%%%%%

\section{Experiments} 
\label{sec:experiments}

In this section, we evaluate the fairness of transaction fraud models trained on public datasets that include demographic information, specifically the Sparkov\footnote{\url{https://www.kaggle.com/datasets/kartik2112/fraud-detection}} and IBMCard\footnote{\url{https://www.kaggle.com/datasets/ealtman2019/credit-card-transactions}} datasets. We employ LightGBM to train these models and utilize group fairness metrics and their normalized versions to evaluate model bias. Our experiments aim to uncover bias patterns across different clusters of fairness metrics discussed in the previous section, focusing on both fraud protection and quality of service metrics.

\subsection{Experimental Setup}

\paragraph{Dataset.} We use the Sparkov dataset~\cite{grover2022fraud}, generated using the Sparkov Data Generation tool\footnote{\url{https://github.com/namebrandon/Sparkov_Data_Generation}}, and the IBMCard dataset~\cite{altman2021synthesizing}, created through a multi-agent virtual world simulation by IBM, as synthetic datasets for fraud detection. The Sparkov dataset includes transactions from 1,000 cardholders with 800 merchants between January 1, 2019, and December 31, 2020, while the IBMCard dataset spans decades and includes transactions from 2,000 US-based cardholders traveling globally. Both datasets contain legitimate and fraudulent transactions and feature transaction-related attributes (transaction time, amount, card identity, merchant identity, and merchant code) as well as demographic attributes (gender, geographical location, and age). For fairness evaluation, gender is used as the sensitive attribute in both datasets.

%\paragraph{Sparkov Dataset.} This synthetic credit card transaction dataset for fraud detection, generated using the Sparkov Data Generation tool\footnote{\url{https://github.com/namebrandon/Sparkov_Data_Generation}}, contains legitimate and fraudulent transactions from 1 Jan 2019 to 31 Dec 2020~\cite{grover2022fraud}. It covers 1,000 cardholders making transactions with a pool of 800 merchants. In addition to transaction-related features (such as transaction time, amount, card identity, merchant identity, and merchant code), the dataset includes demographic attributes such as gender, geographical location, job, and age. In this fairness evaluation, we use gender as the sensitive attribute.

%\paragraph{IBMCard Dataset.} This synthetic dataset comprises over 20 million transactions derived from a multi-agent virtual world simulation conducted by IBM~\cite{altman2021synthesizing}. It includes data from 2000 synthetic consumers based in the United States who travel globally. The dataset spans decades of transactions across multiple cards per consumer and encompasses transaction details like time, amount, card identity, merchant identity, and merchant code. Additionally, it incorporates demographic information such as gender, geographical location, and age. Gender serves as the sensitive attribute in our fairness assessment.

\paragraph{Feature Engineering.} We employ an identical feature engineering pipeline for both datasets. For training the transaction fraud models, we use the behavioral features suggested by~\cite{jha2012employing}.

\paragraph{Training, Validation, and Test Sets.} The datasets for both Sparkov and IBMCard are divided into training, validation, and test sets. For Sparkov, the splits are: (1) a training set of 879K transactions from 675 cardholders, downsampled to 109K transactions with a 4.78\% fraud rate, (2) a validation set of 418K transactions from 308 cardholders, and (3) a test set of 556K transactions from 924 cardholders, ensuring no temporal overlap with the training and validation sets. The Sparkov test dataset shows no significant gender bias, with a normalized true fraud rate parity of 0.024. For IBMCard, the splits are: (1) a training set of 14.7M transactions from 3693 cardholders, downsampled to 376K transactions with a 4.76\% fraud rate, (2) a validation set of 4.8M transactions from 1273 cardholders, and (3) a test set of 4.8M transactions from 1173 cardholders, ensuring no cardholder overlap with the training and validation sets. The IBMCard test dataset exhibits notable gender bias, with a normalized true fraud rate parity of 0.101. 

\paragraph{Training Transaction Fraud Models.} We employ an identical training pipeline for both datasets. We train LightGBM classifiers on the feature-engineered, downsampled training set using cross-entropy as the training objective, while AUC is used as a validation metric~\cite{ke2017lightgbm}. We employ two training strategies: (1) standard empirical risk minimization (ERM), which includes the sensitive attribute (gender) as input during training and inference, and (2) fairness through unawareness, which excludes the sensitive attribute (gender) from the input during training and inference.

\subsection{Fairness Evaluation}

\begin{figure*}[htbp]
    \centering
    \begin{subfigure}[b]{\textwidth}
        \centering
        \includegraphics[width=1.0\textwidth]{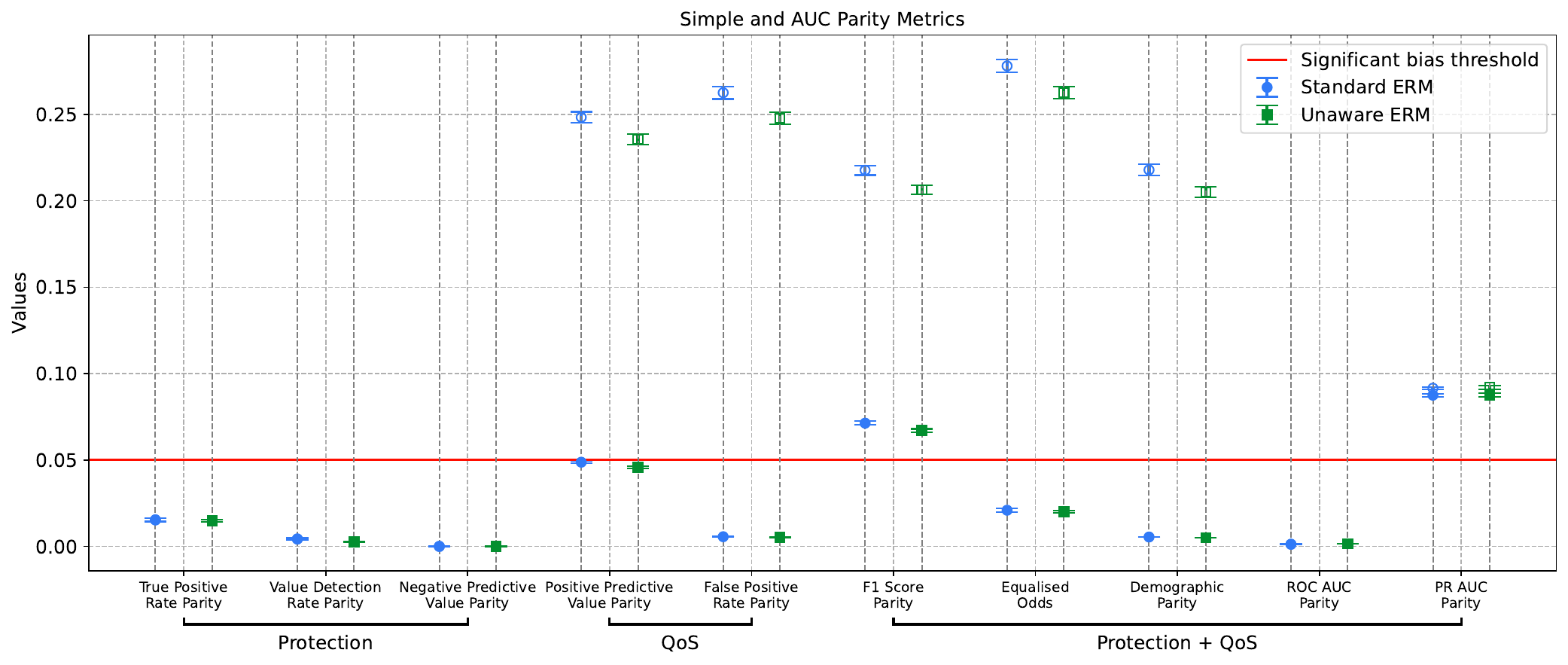}
        \caption{Sparkov dataset}
        \label{fig_sparkov:parity_metrics}
    \end{subfigure}    
    \begin{subfigure}[b]{\textwidth}
        \centering
        \includegraphics[width=1.0\textwidth]{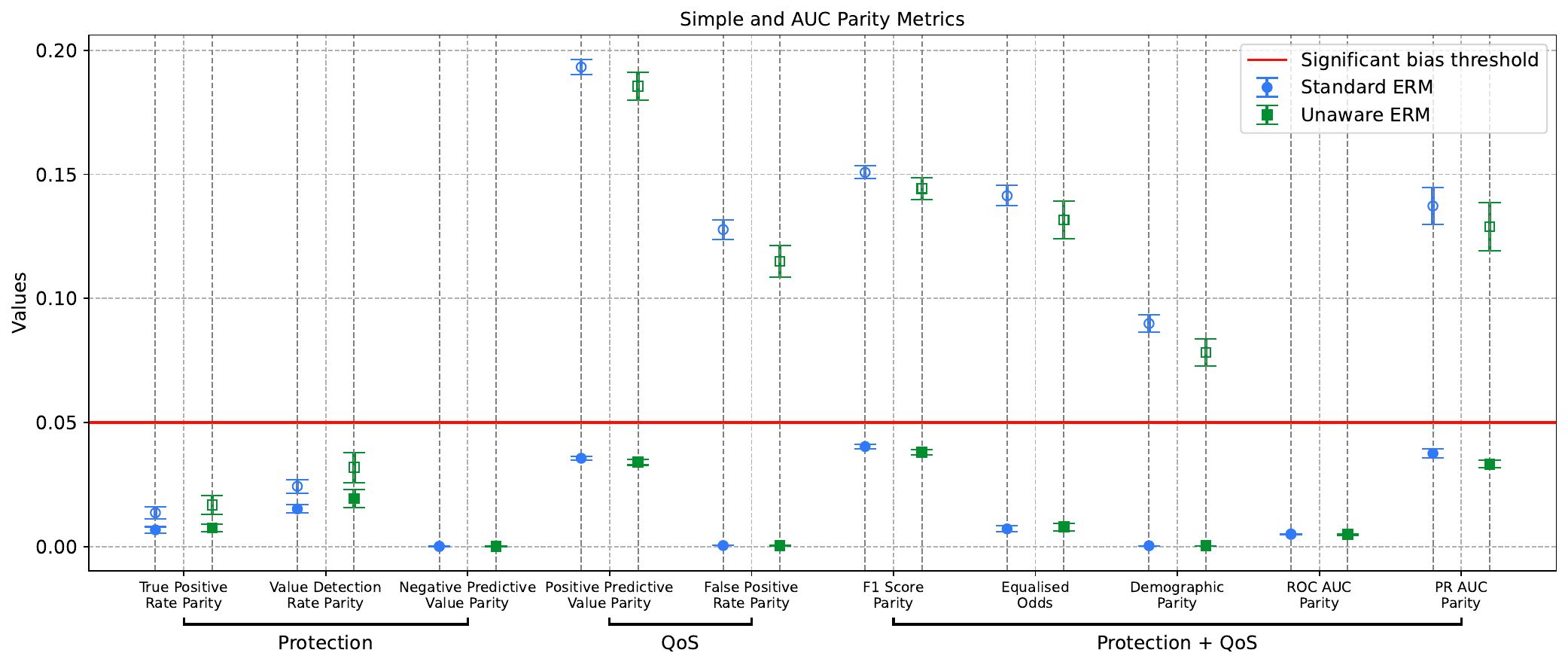}
        \caption{IBMCard dataset}
        \label{fig_ibmcard:parity_metrics}
    \end{subfigure}    
    \caption{Transaction-level fairness evaluation of LightGBM classifier models trained on Sparkov and IBMCard datasets using standard ERM and unaware ERM approaches. Simple parity metrics are computed for a global FP ratio of 5.0. Note that AUC parity metrics are calculated independently of this fixed FP ratio. Each parity metric's normalized value is shown as an unfilled shape (circle/square) along the same vertical line. A significance bias threshold of 0.05 is chosen in accordance with existing literature~\cite{han2023ffb}. For both datasets, significant bias was not observed for protection-related metrics with a global FP ratio of 5.0, even after normalization. However, significant bias was evident for QoS-related metrics with the same global FP ratio after normalization. Similarly, after normalization, significant bias was observed for combined protection and QoS-related metrics, except for ROC AUC parity.}
    \Description{Transaction-level simple and AUC parity metrics.}
    \label{fig:parity_metrics}
\end{figure*}

\begin{figure*}[htbp]
    \centering
    % First row of subfigures
    \begin{subfigure}[b]{0.47\textwidth}
        \centering
        \includegraphics[width=\textwidth]{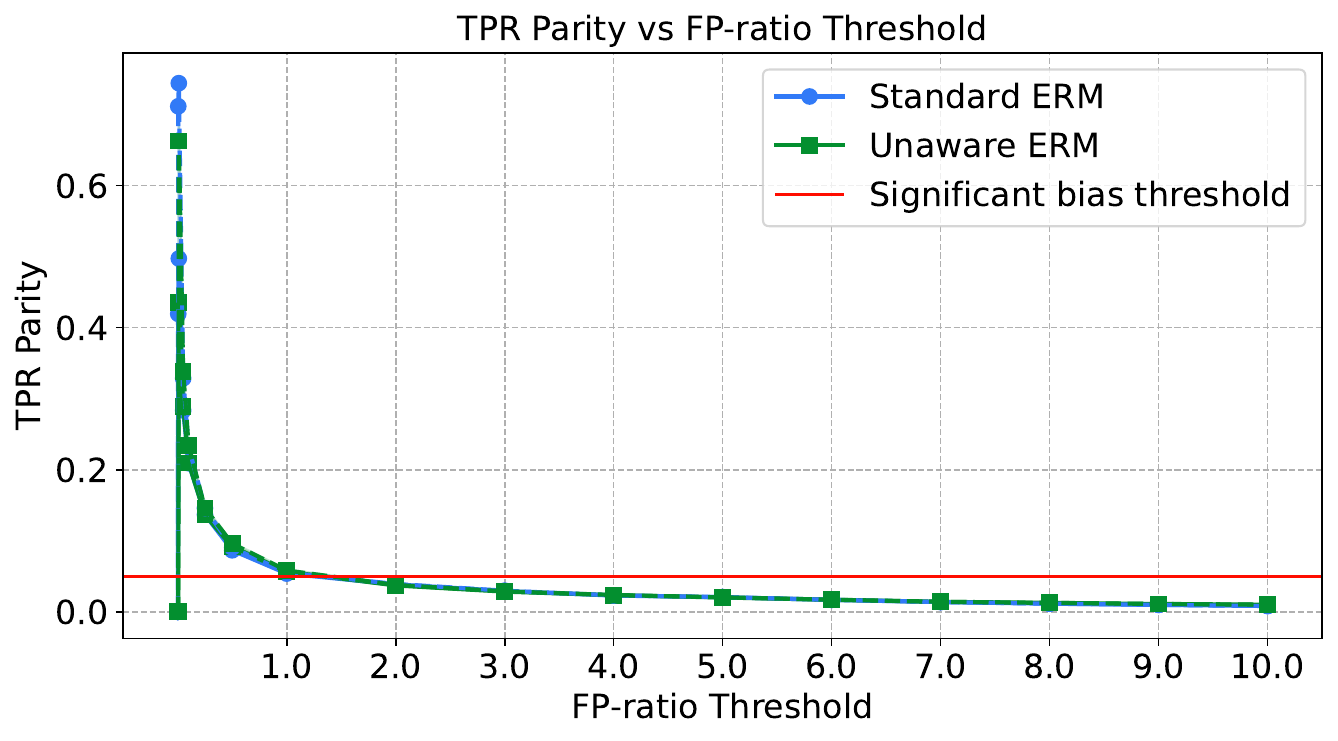}
        \label{fig_sparkov:tpr_parity_metrics_at_fp_ratio_threshold}
    \end{subfigure}
    \hfill
    \begin{subfigure}[b]{0.47\textwidth}
        \centering
        \includegraphics[width=\textwidth]{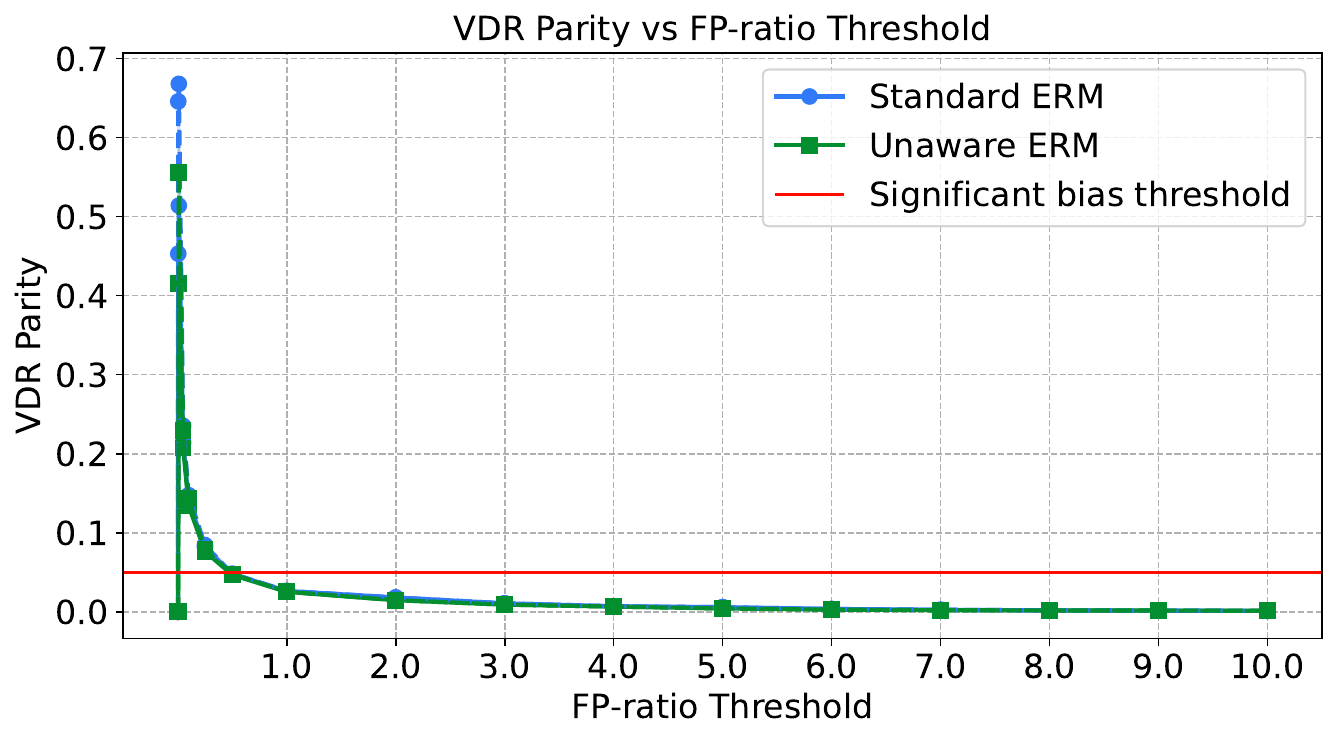}
        \label{fig_sparkov:vdr_parity_metrics_at_fp_ratio_threshold}
    \end{subfigure}
    \vspace{-0.5cm}
    \begin{center}
        \subcaption{Sparkov dataset}
    \end{center}
    % Second row of subfigures
    \begin{subfigure}[b]{0.47\textwidth}
        \centering
        \includegraphics[width=\textwidth]{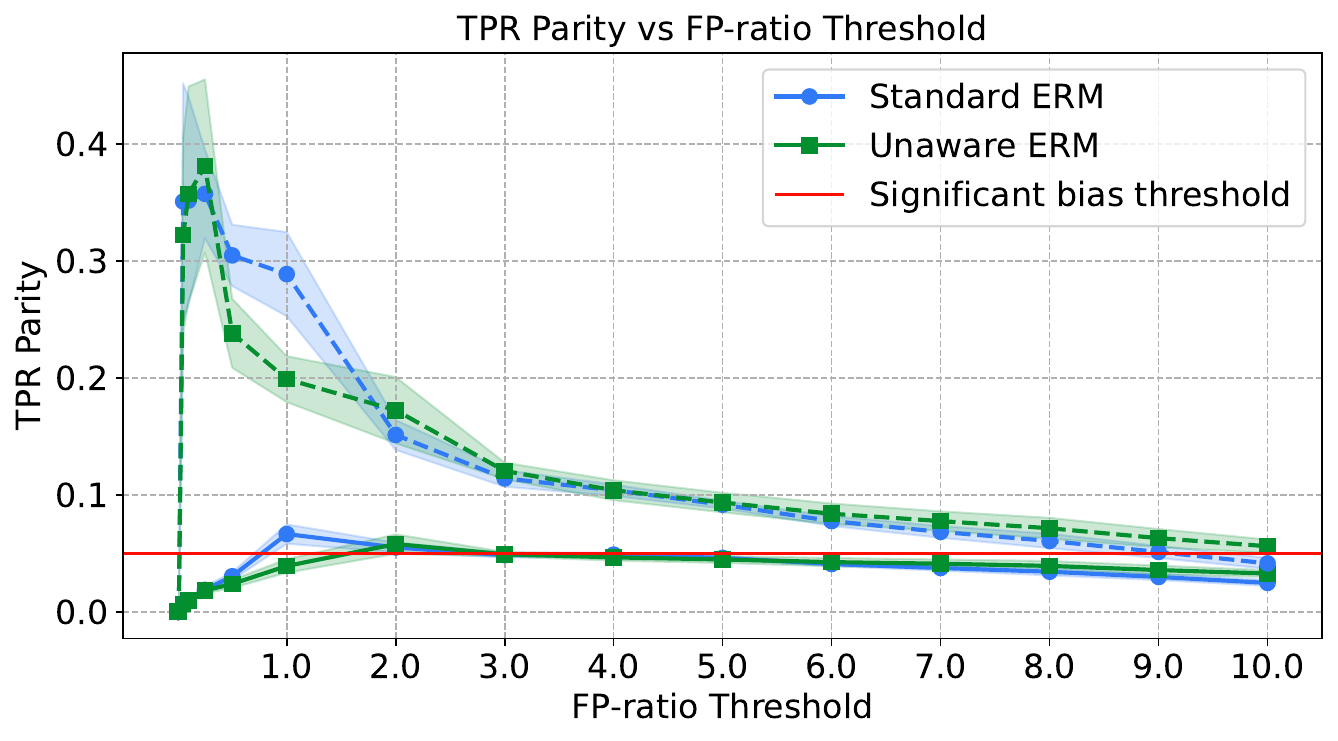}
        \label{fig_ibmcard:tpr_parity_metrics_at_fp_ratio_threshold}
    \end{subfigure}
    \hfill
    \begin{subfigure}[b]{0.47\textwidth}
        \centering
        \includegraphics[width=\textwidth]{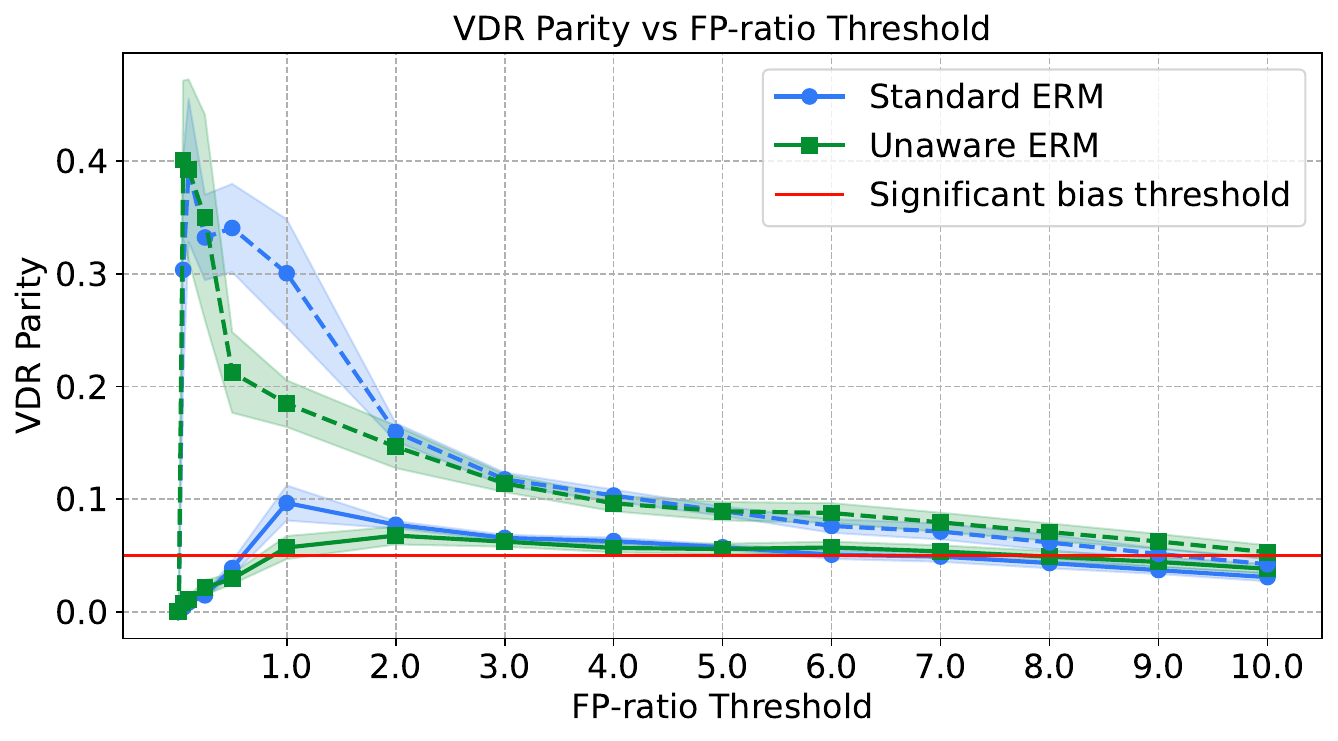}
        \label{fig_ibmcard:vdr_parity_metrics_at_fp_ratio_threshold}
    \end{subfigure}
    \vspace{-0.5cm}
    \begin{center}
        \subcaption{IBMCard dataset}
    \end{center}
    \vspace{-0.5cm}
    \caption{Transaction-level fairness evaluation of LightGBM classifier models trained on Sparkov and IBMCard datasets using standard ERM and unaware ERM approaches. The figure shows TPR Parity, VDR Parity, and their normalized values (dashed lines) at various group-wise FP ratio thresholds. A significance bias threshold of 0.05 is chosen in accordance with existing literature~\cite{han2023ffb}. Significant bias in TPR and VDR parity was observed at lower group-wise FP ratio thresholds (high-precision regime) for both datasets. As the group-wise FP ratio decreases substantially, the TPR and VDR performance for both genders deteriorates, potentially causing the observed bias to diminish.}
    \Description{Transaction-level TPR and VDR parity metrics at various group-wise FP-ratio thresholds.}
    \label{fig:parity_metrics_at_fp_ratio_threshold}
\end{figure*}

\paragraph{Transaction-level Fairness Evaluation.} We present the transaction-level fairness evaluation results in Figures~\ref{fig:parity_metrics}~and~\ref{fig:parity_metrics_at_fp_ratio_threshold}, averaged over ten experimental runs using different random states for training the LGBMClassifier. In Figure~\ref{fig:parity_metrics}, the simple metrics are computed for a fixed global FP ratio of 5.0. First, we identify the corresponding global score threshold. Then, using this score threshold, we compute the group-wise utility metrics (such as TPR and FPR) for each group. Finally, we calculate the parity values from these group-wise utility metrics. In Figure~\ref{fig:parity_metrics_at_fp_ratio_threshold}, for each group-wise FP-ratio, we first identify the corresponding group-wise score thresholds for each group. Next, we compute the group-wise utility metrics (TPR and VDR) using these score thresholds. We then calculate the parity values from these group-wise utility metrics. Here, we evaluate the fraud protection parity for the same group-wise quality of service (same FP-ratio for both groups).

\paragraph{Key Findings from Experiments.} Our experiments on the Sparkov and IBMCard datasets using LightGBM models revealed several critical insights: 
\begin{itemize}
\item Protection-related Metrics: No significant bias was observed in protection-related metrics with a global FP ratio of 5.0, even after normalization. Very low NPV parity can be attributed to the model's higher accuracy in predicting non-fraudulent transactions for both genders, given the ample number of non-fraudulent transactions in the dataset.
\item Quality of Service (QoS) related Metrics: Significant bias emerged after normalization for QoS-related metrics with the same FP ratio. This indicates that normalized metrics are crucial for detecting biases that might go unnoticed.
\item Combined Metrics: Significant bias was observed in combined protection and QoS-related metrics post-normalization, except for ROC AUC parity. The very low ROC AUC parity is due to the high ROC AUC values (above 0.9) in our experiments. Unlike PR AUC, the ROC AUC is a less suitable metric for highly imbalanced classification, focusing more on the majority class.
\item High-Precision Regime: Significant TPR and VDR parity bias was observed at smaller group-wise FP ratios. This indicates that as the rate of false rejections decreases, the level of fraud protection afforded to each gender varies significantly, highlighting a substantial bias. As the FP ratio decreases substantially, the TPR and VDR performance for both genders deteriorates, potentially causing the observed bias to diminish or become less apparent.
\item Impact of Measurement Approach: For the IBMCard dataset, significant bias in normalized TPR and VDR parity metrics was not observed at a global FP ratio of 5.0 (Figure~\ref{fig:parity_metrics}), but was evident at a group-wise FP ratio of 5.0 (Figure~\ref{fig:parity_metrics_at_fp_ratio_threshold}). This discrepancy highlights that the observed bias can depend on the measurement approach, underscoring the need to consider both global and group-wise FP ratios in bias analysis.
\item Fairness Through Unawareness: This approach did not help mitigate bias, indicating the presence of correlated proxies for gender within the dataset.
\end{itemize}
These key findings are consistent across both datasets despite their synthetic nature and differences in data generation processes.

% !TEX root =  main.tex
%%%%%%%%%%%%%%%%%%%%%%%%%%%%%%%%%%%%%
%%%%%%%%%%%%%%%%%%%%%%%%%%%%%%%%%%%%%

\section{Socio-Technical Fairness in Transaction Fraud Models} 
\label{sec:socio-tech-fairness}

In this section, we discuss the challenges and approaches for integrating socio-technical fairness into transaction fraud models. We emphasize the importance of considering the lived experiences of individuals affected by these models, balancing the objectives of different stakeholders, and evaluating fairness at the cardholder level. These elements are crucial for developing a more comprehensive and equitable approach to fairness in transaction fraud models.
%, and ensuring continuous monitoring

\subsection{Incorporating Lived Experiences}

Algorithmic fairness in machine learning models involves abstracting and simplifying complex social scenarios into standard mathematical objectives. While this approach facilitates the development of fairness-aware algorithms, it can be misused when implemented without considering specific circumstances and contexts. This context-agnostic method often overlooks the lived experiences of individuals, particularly those from historically marginalized groups, whose interactions with AI systems can vary significantly~\cite{selbst2019fairness}.

In the context of transaction fraud models, the impact of wrongful flagging varies. For example, incorrectly flagging a transaction for essential goods can have immediate and critical consequences for cardholders, whereas false positives on luxury items, while inconvenient, generally have less severe effects. Addressing these issues requires a human-centric perspective, emphasizing the importance of including affected individuals and communities in defining and implementing AI fairness~\cite{johnson2017ai}.

\subsection{Balancing Stakeholder Objectives}

Understanding the distinct challenges in formalizing fairness for transaction fraud models is crucial. Unlike lending practices, where biases across demographic groups are well-documented and legislated against~\cite{williams_changing_2005,Home_Mortgage,Discrimination}, fraud detection requires balancing fraud protection with service quality. Unfairly declined transactions can severely impact users, necessitating a nuanced approach to fairness that considers both protection against fraud and maintaining quality service.

Stakeholders often have conflicting goals that cannot be easily resolved through increased data collection or modeling adjustments~\cite{eubanks2018automating,o2020near}. Cardholders prioritize avoiding false positives on essential transactions, while financial institutions aim to minimize fraud liability, which involves accurately identifying high-value fraudulent transactions. Navigating these conflicts requires careful consideration of trade-offs in stakeholders' fairness objectives and determining whose objectives should be prioritized.

\subsection{Cardholder-Level Evaluation}

Fairness evaluation in transaction fraud models should occur at both the transaction and cardholder levels, considering evolving fraud patterns and the feedback of affected individuals. While transaction-level fairness ensures each transaction is treated impartially, cardholder-level fairness provides a holistic view of bias by evaluating the cumulative experience of cardholders over time.

A weighted aggregate of transaction-level evaluations aims to provide a fair cumulative experience for cardholders, e.g., evaluating the fairness of transaction fraud models based on each cardholder's last 100 transactions. This comprehensive evaluation uncovers systemic biases that might be overlooked when considering only individual transactions. Determining the appropriate weighting of transaction-level assessments to achieve fairness across all cardholders is challenging, as socio-technical aspects like transaction frequency, types, and context play a crucial role in ensuring accurate fairness metrics.

\section{Conclusion} 
\label{sec:conclusion}

This paper presents a comprehensive evaluation of fairness in transaction fraud models, pioneering the algorithmic bias audit in this domain. Our contributions are substantial:
\begin{enumerate}
\item First Published Bias Audit: We conducted the first algorithmic bias audit of transaction fraud models, shedding light on previously unexplored aspects of fairness in this context.
\item Clustering of Fairness Metrics: By categorizing fairness metrics based on their objectives--either ensuring equal protection against fraud or delivering uniform quality of service--we uncovered distinct patterns of bias in these clusters.
\item Socio-technical Challenges: Our discussion highlighted the complexities of selecting appropriate fairness metrics and bias mitigation strategies, emphasizing the need to consider the nature of transactions and socio-technical factors.
\end{enumerate}
These contributions provide a comprehensive framework for evaluating and improving fairness in transaction fraud models. In particular, our experiments using LightGBM models on the Sparkov and IBMCard datasets revealed critical insights. Significant bias was absent in protection-related metrics with a global FP ratio of 5.0, even after normalization. However, normalization exposed substantial bias in quality of service (QoS) metrics and combined metrics, except for ROC AUC parity. Significant TPR and VDR parity bias were observed in high-precision regimes at smaller group-wise FP ratios. The Fairness Through Unawareness approach did not help mitigate bias, indicating correlated proxies for gender. 

Our study is limited by the use of synthetic datasets, which may not fully capture the complexities of real-world transaction data. Nevertheless, despite the synthetic nature and variations in data generation processes, our key findings remained consistent, underscoring the robustness and generalizability of our results. Additionally, our focus on gender as the sensitive attribute leaves room for further exploration of intersectional fairness involving multiple demographic attributes.

Future research should build on these insights by integrating socio-technical factors, particularly through cardholder-level fairness evaluations that weight transactions based on their context. This approach will offer a deeper understanding of fairness by addressing the varying nature of transactions. Additionally, developing bias mitigation strategies tailored to the unique challenges of transaction fraud models--such as class imbalance, the importance of transaction value, and the sequential nature of data--is crucial. Advancing fair synthetic data generation methods and employing feature attribution algorithms will enhance our ability to identify and address proxy features of demographic attributes, thereby supporting more effective bias mitigation. These efforts will be pivotal in ensuring fairness in transaction fraud models, paving the way for more equitable and effective systems.

\begin{acks}
We thank Featurespace for the financial support and resources offered during the completion of this research.
\end{acks}

\bibliographystyle{ACM-Reference-Format}
\bibliography{main}

\clearpage
\onecolumn
\appendix
% !TEX root =  main.tex
%%%%%%%%%%%%%%%%%%%%%%%%%%%%%%%%%%%%%
%%%%%%%%%%%%%%%%%%%%%%%%%%%%%%%%%%%%%

\section{Additional Details} 
\label{app:details}

\subsection{Additional Details: Algorithmic Fairness in Transaction Fraud Models}

\paragraph{Fairness Metrics.} Transaction fraud prediction is a binary classification problem aimed at determining whether a transaction is fraudulent based on its history. The dataset's feature set is denoted as $X$,  and the class label as $Y$. Additionally, we consider a binary sensitive attribute $G \in \{ \textnormal{Male}, \textnormal{Female} \}$ associated with each example. The group fairness metrics are calculated using the values from the confusion matrix (true positives (\texttt{TP}), false positives (\texttt{FP}), true negatives (\texttt{TN}), and false negatives (\texttt{FN})). \texttt{TP} represents correctly identifying a fraudulent transaction, while \texttt{FP} denotes a legitimate transaction wrongly flagged as fraudulent and subsequently declined. \texttt{FN} occurs when a fraudulent transaction successfully proceeds, leading to financial losses for the financial institution or customer. \texttt{TN} refers to correctly identifying a legitimate transaction.

\paragraph{Normalized Parity Metrics.} For all fairness metrics discussed in Section~\ref{subsec:metrics}, we also compute their normalized variants. For example, we compute the normalized FPR parity as follows:
\[
\frac{\abs{\texttt{FPR}_{G=\textnormal{Male}} - \texttt{FPR}_{G=\textnormal{Female}}}}{\max \bc{\texttt{FPR}_{G=\textnormal{Male}}, \texttt{FPR}_{G=\textnormal{Female}}}} .
\]
This normalization is essential for identifying bias due to the class imbalance inherent in the transaction fraud detection problem. We chose this normalization because it works very well when a utility metric takes a very small value, which is the case with some of our metrics due to class imbalance. This normalization essentially scales the parity values to the range $[0, 1]$ (since the utility metrics take non-negative values), where 1 indicates the maximum level of bias achievable. For instance, in our experiments in Section~\ref{sec:experiments}, significant biases in important metrics such as Precision parity, FPR parity, and Equalized Odds go unnoticed without this normalization.

\paragraph{Bias Threshold.} The ideal value for all the above fairness metrics is 0. A significance bias threshold of 0.05 is chosen in accordance with existing literature~\cite{han2023ffb}. This threshold may vary depending on the context and the specific metric used.

\subsection{Additional Details: Socio-Technical Fairness in Transaction Fraud Models}

\paragraph{Continuous Monitoring.} Establishing mechanisms for cardholders to provide feedback on flagged transactions and contest decisions is vital. Transparent communication about the system's operation and the rationale behind flagged transactions fosters user trust and ensures perceived fairness~\cite{explain,disenta}. As fraud patterns evolve, regular updates and evaluations are necessary to maintain both accuracy and fairness. Ongoing assessments must account for data drift, model drift, and concept drift~\cite{drift,drift1}.

\subsection{Additional Details: Experiments}

\paragraph{Dataset.} Table~\ref{table:dataset_stats} summarizes the statistics for both Sparkov and IBMCard datasets.

\begin{table*}[ht]
\centering
\caption{Summary of statistics for the Sparkov and IBMCard datasets (train, validation, test, and transaction-level downsampled training set). Both datasets are reasonably balanced in terms of the number of transactions across genders. The Sparkov test dataset shows no significant gender bias in normalized true fraud rate parity (0.024), while the IBMCard test dataset exhibits notable gender bias (0.101). There may be correlated features related to the sensitive attribute in these datasets.}
\label{table:dataset_stats}
\begin{tabular}{lllllllll}
\toprule
& \multicolumn{4}{c}{Sparkov} & \multicolumn{4}{c}{IBMCard} \\ 
\cmidrule(lr){2-5} \cmidrule(lr){6-9}
                      & Train & Valid & Test & Down & Train & Valid & Test & Down \\ \midrule
\# cards              & 675   & 308   & 924  & 675  & 3693  & 1273  & 1173 & 3438 \\
\# cards (Male)       & 336   & 150   & 456  & 336  & 1854  & 601   & 549  & 1718 \\
\# cards (Female)     & 339   & 158   & 468  & 339  & 1839  & 672   & 624  & 1720 \\ \addlinespace
\# defrauded cards    & 523   & 239   & 218  & 523  & 1641  & 566   & 545  & 1641 \\
\# defrauded cards (Male) & 265 & 117 & 102  & 265  & 797   & 270   & 257  & 797  \\
\# defrauded cards (Female) & 258 & 122 & 116 & 258 & 844 & 296 & 288 & 844  \\ \addlinespace
\# transactions       & 879K  & 418K  & 556K & 109K & 14.7M & 4.8M  & 4.8M & 376K \\
\# transactions (Male) & 397K & 190K  & 251K & 50K  & 7.2M  & 2.4M  & 2.2M & 185K \\
\# transactions (Female) & 482K & 228K & 305K & 59K & 7.5M & 2.4M & 2.6M & 191K \\ \addlinespace
fraud rate           & 0.59  & 0.55  & 0.39 & 4.78 & 0.12  & 0.12  & 0.12 & 4.76 \\
fraud rate (Male)    & 0.67  & 0.59  & 0.39 & 5.35 & 0.12  & 0.12  & 0.11 & 4.76 \\
fraud rate (Female)  & 0.53  & 0.52  & 0.39 & 4.31 & 0.12  & 0.13  & 0.13 & 4.76 \\ \bottomrule
\end{tabular}
\end{table*}

\paragraph{Feature Engineering.} For training the transaction fraud models, we use the following features as suggested by~\cite{jha2012employing}: raw features from the dataset (merchant code, merchant identity, transaction amount, and gender), features computed using transaction time (transaction hour, transaction day of the week, and transaction month), cardholder-related features, cardholder and merchant code interaction features, and cardholder and merchant identity interaction features. The cardholder-related and interaction features include running-window aggregations (such as the count of transactions and the sum of transaction value) across cardholder transactions, capturing cardholder spending patterns. For the fairness through unawareness approach, we simply remove the gender feature.

\paragraph{Pre-processing.} We apply target encoding to the categorical features "merchant code" and "merchant identity." The target label 0 corresponds to non-fraudulent transactions, while 1 corresponds to fraudulent transactions. For the sensitive attribute "gender," we use binary encoding (Male: 0, Female: 1). Finally, we use a standard scaler to normalize the features.

\paragraph{Hyperparameter Tuning.} Hyperparameters are tuned using Optuna~\cite{akiba2019optuna} (see Table~\ref{table:hyperparameters}).

\begin{table*}[ht]
\centering
\caption{Hyperparameters used to train LightGBM classifier models on the Sparkov and IBMCard datasets.}
\label{table:hyperparameters}
\begin{tabular}{lll}
\toprule
\textbf{Parameters} & \textbf{Sparkov} & \textbf{IBMCard} \\ \midrule
\texttt{objective} & binary & binary \\ 
\texttt{metric} & auc & auc \\ 
\texttt{is\_unbalance} & True & True \\ 
\texttt{boosting\_type} & gbdt & gbdt \\ 
\texttt{learning\_rate} & 0.095 & 0.05 \\ 
\texttt{colsample\_bytree} & 0.65 & 0.8 \\ 
\texttt{subsample} & 0.95 & 0.8 \\ 
\texttt{feature\_fraction} & 0.7 & 0.8 \\ 
\texttt{max\_bin} & 200 & 63 \\ 
\texttt{num\_leaves} & 100 & 225 \\ 
\texttt{max\_depth} & 50 & 15 \\ 
\texttt{n\_estimators} & 400 & 500 \\ 
\texttt{num\_iterations} & 400 & 1000 \\ 
\texttt{early\_stopping\_round} & 90 & 50 \\ 
\texttt{early\_stopping\_min\_delta} & 0.0001 & 0.0005 \\ 
\texttt{lambda\_l1} & 0.0 & 0.2 \\ 
\texttt{lambda\_l2} & 2.5 & 0.0 \\ \bottomrule
\end{tabular}
\end{table*}

\paragraph{Group-wise FP Ratio Parity.} In Figure~\ref{fig:parity_metrics}, for a global FP ratio of 5.0, we observe the group-wise FP ratio parity, as summarized in Table~\ref{table:fp_ratio_parity}.

\begin{table}[ht]
\centering
\caption{Transaction-level fairness evaluation of LightGBM classifier models trained on Sparkov and IBMCard datasets using standard ERM and unaware ERM approaches. The group-wise FP ratio parity is computed for a global FP ratio of 5.0. The parity metric's normalized value is presented in the bracket. Significant bias values are bolded. A significance bias threshold of 0.05 is chosen in accordance with existing literature~\cite{han2023ffb}.}
\label{table:fp_ratio_parity}
\begin{tabular}{lll}
\toprule
\textbf{Dataset} & \textbf{Standard ERM} & \textbf{Unaware ERM} \\
\hline
Sparkov & \textbf{1.683}$_{\pm0.024}$ (\textbf{0.291}$_{\pm0.004}$) & \textbf{1.586}$_{\pm0.022}$ (\textbf{0.277}$_{\pm0.003}$) \\
IBMCard & \textbf{1.292}$_{\pm0.028}$ (\textbf{0.225}$_{\pm0.004}$) & \textbf{1.236}$_{\pm0.042}$ (\textbf{0.217}$_{\pm0.007}$) \\
\bottomrule
\end{tabular}
\end{table}

\begin{figure*}[htbp]
    \centering
    \begin{subfigure}[b]{\textwidth}
        \centering
        \includegraphics[width=1.0\textwidth]{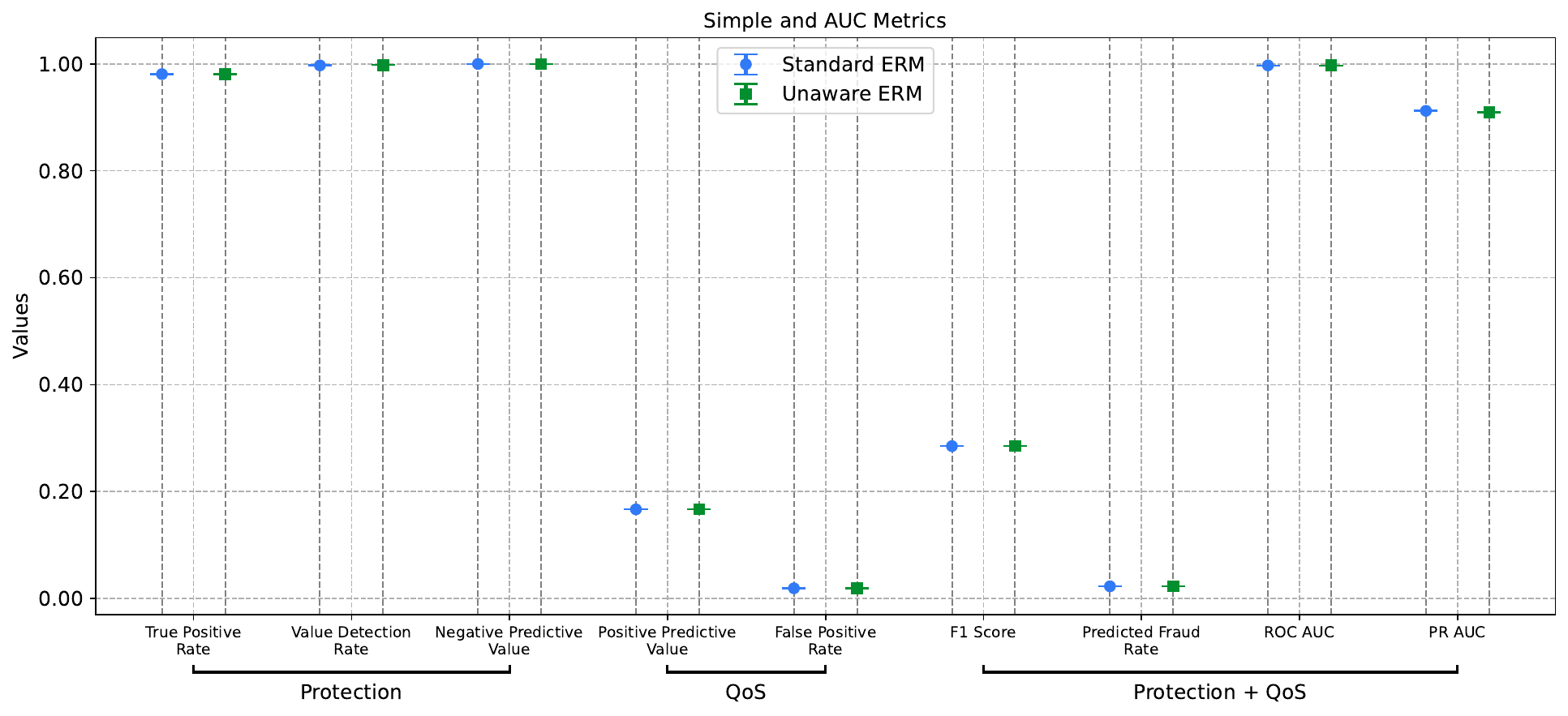}
        \caption{Sparkov dataset}
        \label{fig_sparkov:utility_metrics}
    \end{subfigure}    
    %\vspace{2cm} 
    \begin{subfigure}[b]{\textwidth}
        \centering
        \includegraphics[width=1.0\textwidth]{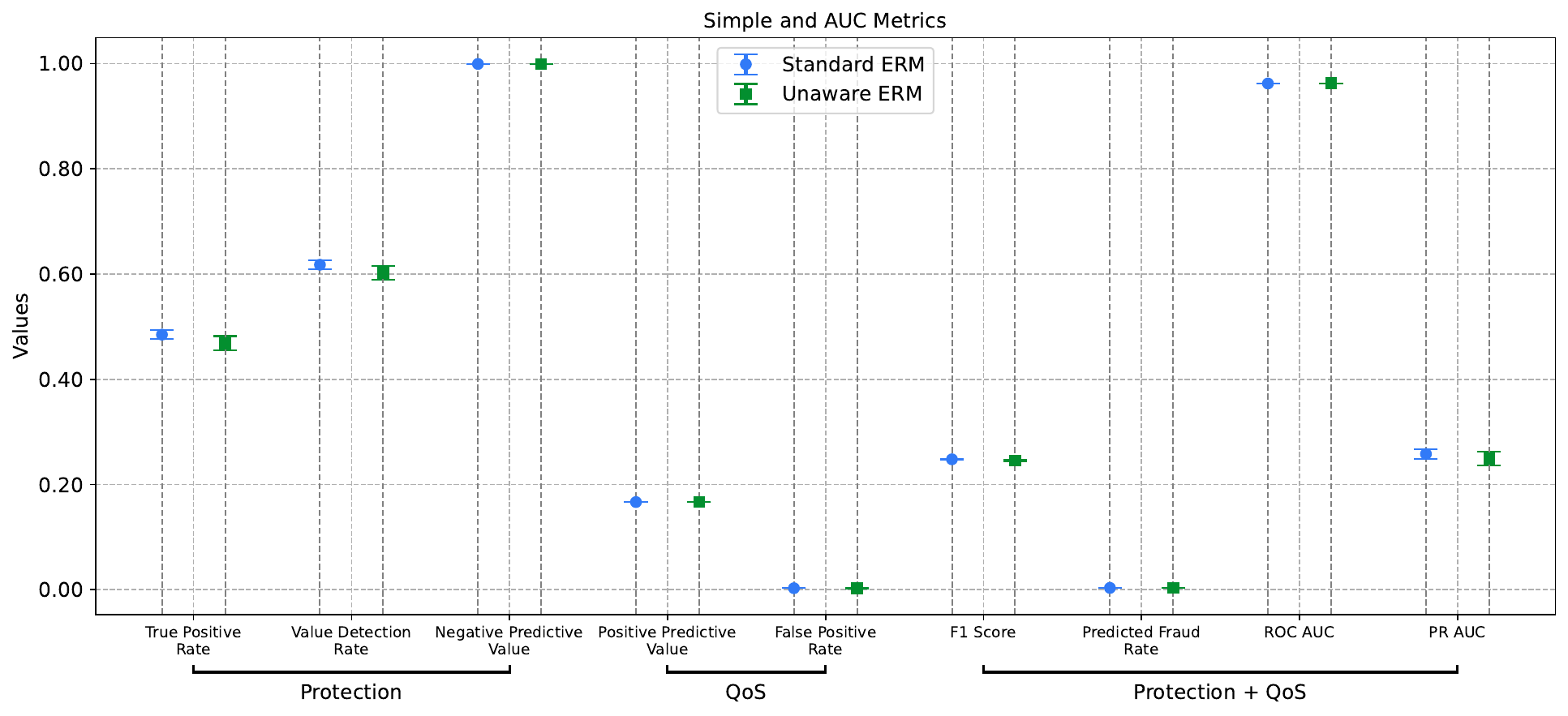}
        \caption{IBMCard dataset}
        \label{fig_ibmcard:utility_metrics}
    \end{subfigure}    
    \caption{Transaction-level utility evaluation of LightGBM classifier models trained on Sparkov and IBMCard datasets using standard ERM and unaware ERM approaches. Simple metrics are computed for a global FP ratio of 5.0. Note that AUC metrics are calculated independently of this fixed FP ratio.}
    \Description{Transaction-level simple and AUC metrics.}
    \label{fig:utility_metrics}
\end{figure*}

\begin{figure*}[htbp]
    \centering
    % First row of subfigures
    \begin{subfigure}[b]{0.47\textwidth}
        \centering
        \includegraphics[width=\textwidth]{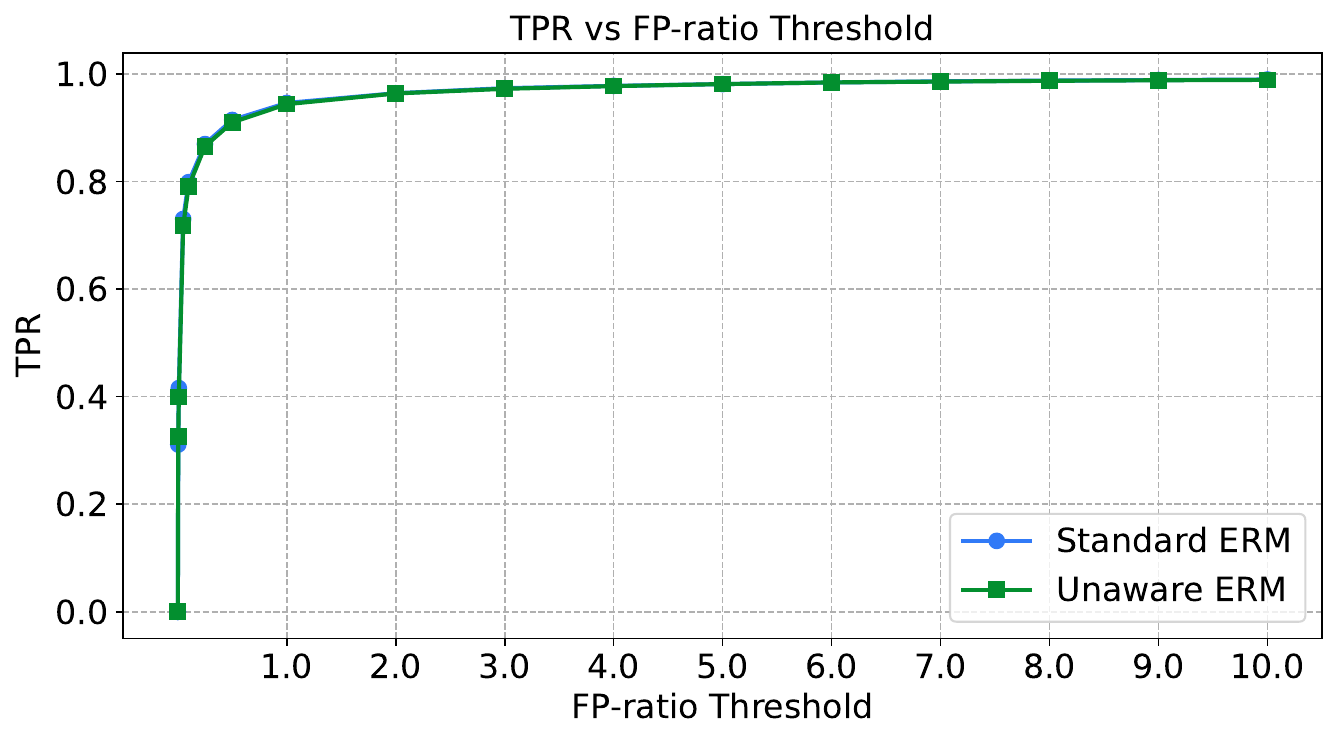}
        \label{fig_sparkov:tpr_metrics_at_fp_ratio_threshold}
    \end{subfigure}
    \hfill
    \begin{subfigure}[b]{0.47\textwidth}
        \centering
        \includegraphics[width=\textwidth]{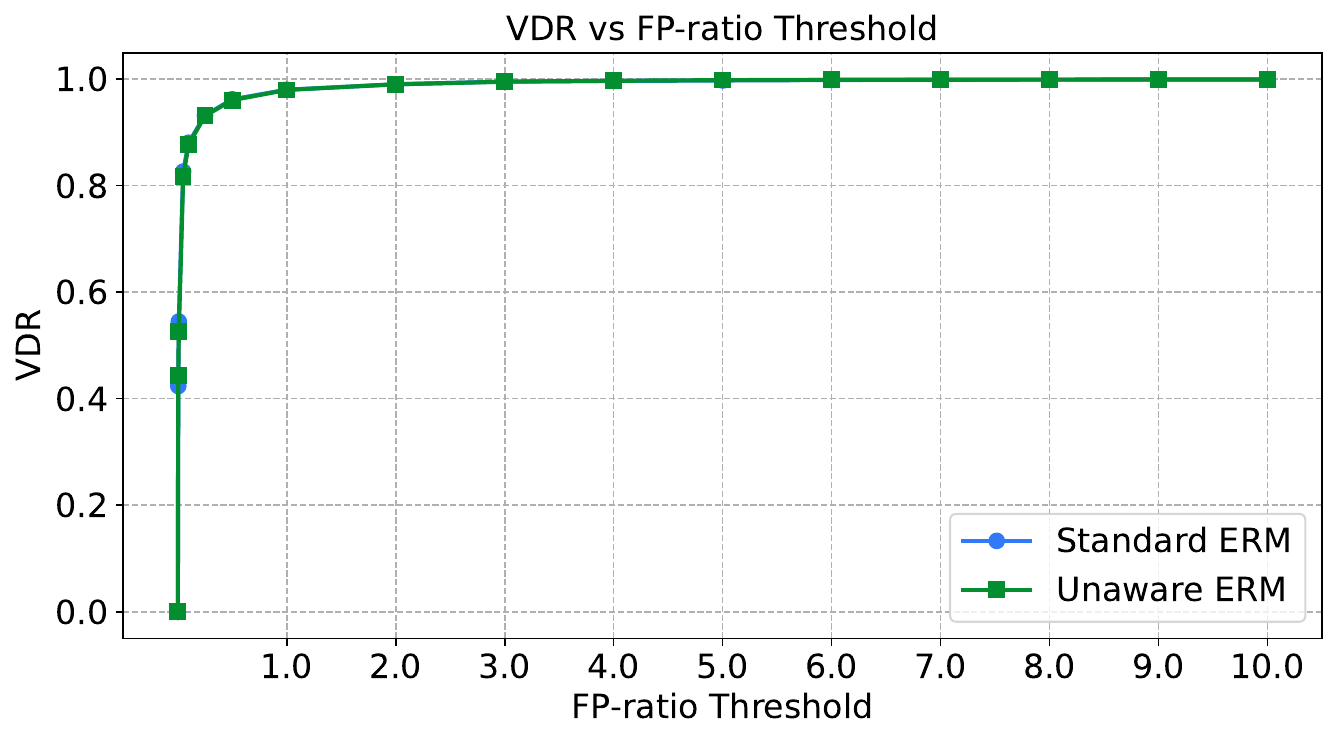}
        \label{fig_sparkov:vdr_metrics_at_fp_ratio_threshold}
    \end{subfigure}
    \vspace{-0.5cm}
    \begin{center}
        \subcaption{Sparkov dataset}
    \end{center}
    % Second row of subfigures
    \begin{subfigure}[b]{0.47\textwidth}
        \centering
        \includegraphics[width=\textwidth]{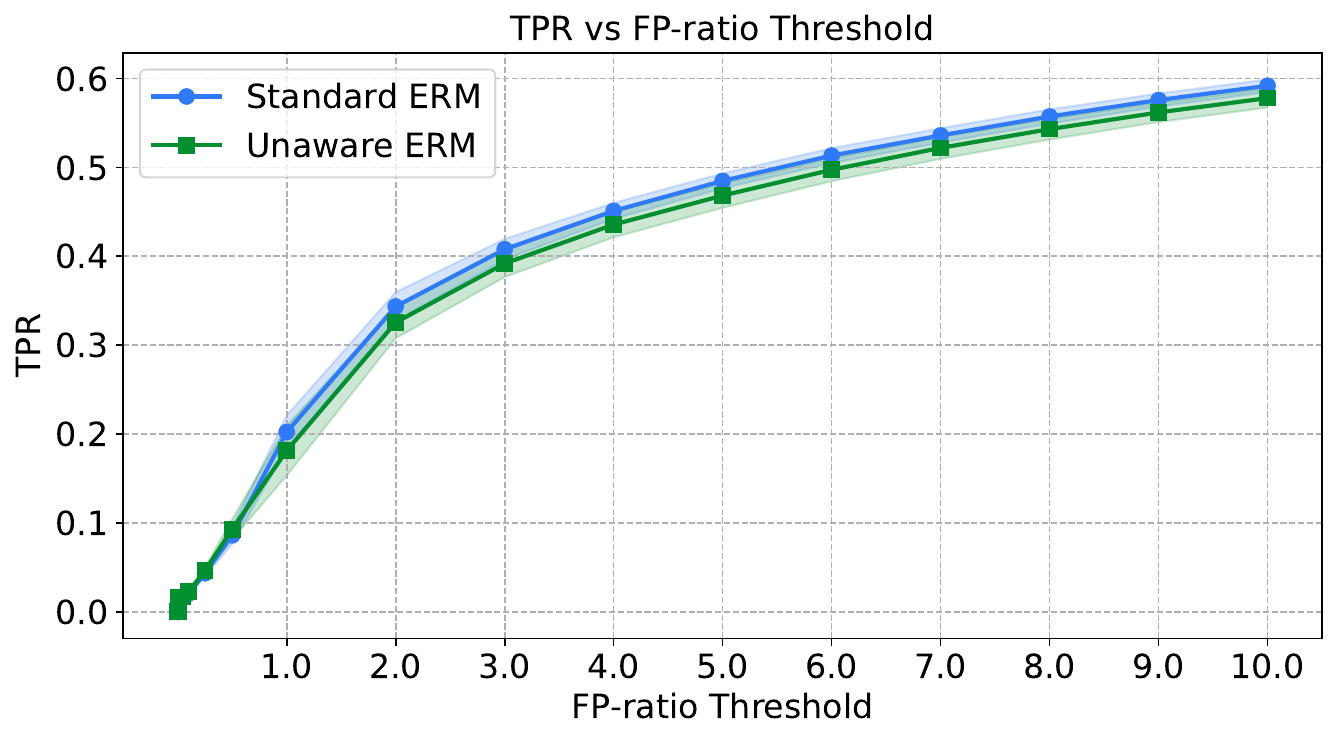}
        \label{fig_ibmcard:tpr_metrics_at_fp_ratio_threshold}
    \end{subfigure}
    \hfill
    \begin{subfigure}[b]{0.47\textwidth}
        \centering
        \includegraphics[width=\textwidth]{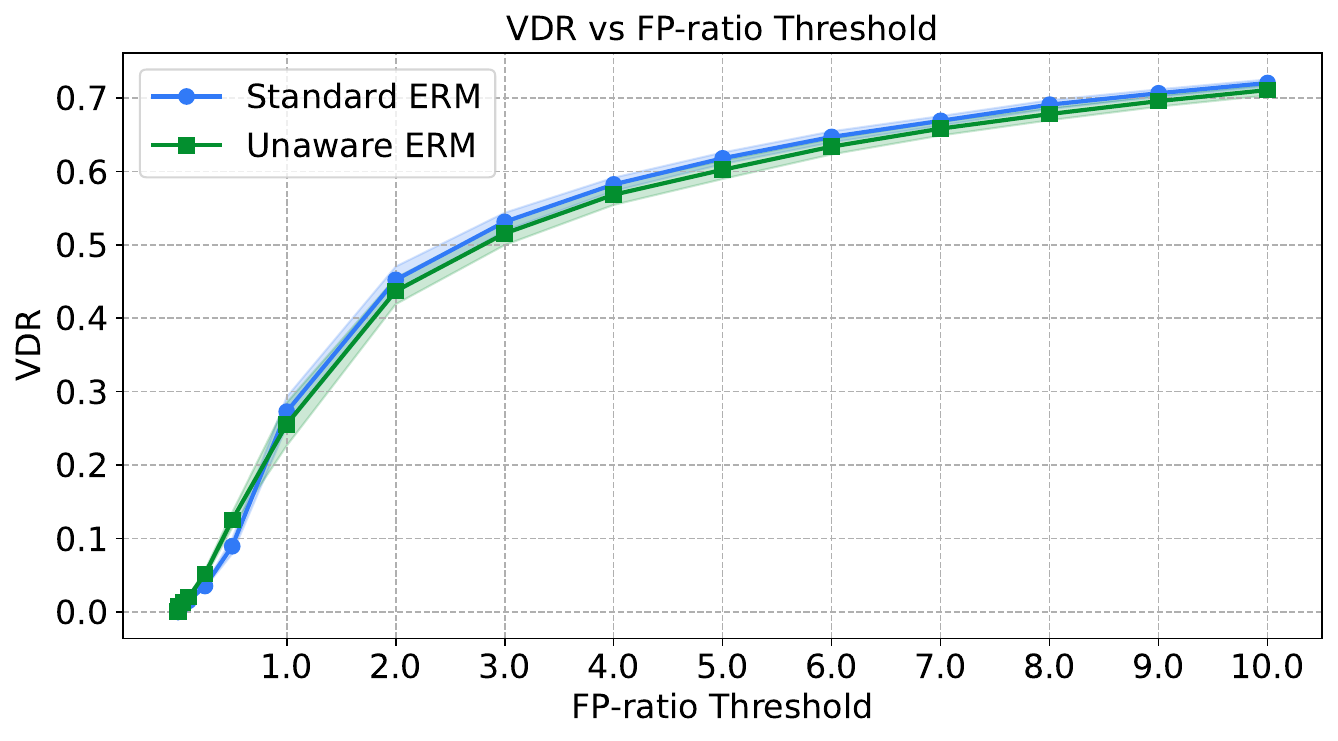}
        \label{fig_ibmcard:vdr_metrics_at_fp_ratio_threshold}
    \end{subfigure}
    \vspace{-0.5cm}
    \begin{center}
        \subcaption{IBMCard dataset}
    \end{center}
    \vspace{-0.5cm}
    \caption{Transaction-level utility evaluation of LightGBM classifier models trained on Sparkov and IBMCard datasets using standard ERM and unaware ERM approaches. The figure shows TPR and VDR at various global FP-ratio thresholds.}
    \Description{Transaction-level TPR and VDR metrics at various global FP-ratio thresholds.}
    \label{fig:utility_metrics_at_fp_ratio_threshold}
\end{figure*}

\paragraph{Transaction-level Utility Evaluation.} We present the transaction-level utility evaluation results in Figures~\ref{fig:utility_metrics}~and~\ref{fig:utility_metrics_at_fp_ratio_threshold}, averaged over ten experimental runs using different random states for training the LGBMClassifier. In Figure~\ref{fig:utility_metrics}, the simple metrics are computed for a fixed global FP ratio of 5.0. First, we identify the corresponding global score threshold. Then, using this score threshold, we compute the global utility metrics (such as TPR and FPR) for each group. In Figure~\ref{fig:utility_metrics_at_fp_ratio_threshold}, for each global FP-ratio threshold, we first identify the corresponding global score threshold. Next, we compute the global utility metrics (TPR and VDR) using these score thresholds.

\end{document}